\documentclass[10pt,twocolumn,letterpaper]{article}

\usepackage{iccv}
\usepackage{times}
\usepackage{epsfig}
\usepackage{graphicx}
\usepackage{amsmath}
\usepackage{amssymb}
\usepackage{multirow}
\usepackage{wrapfig}
\usepackage[accsupp]{axessibility}

\newcommand{\op}[1]{\operatorname{#1}}
\newcommand{\mbf}[1]{\mathbf{#1}}

\newcommand{\app}{\raise.17ex\hbox{$\scriptstyle\sim$}}

\def\x{$\times$}

\newlength\savewidth\newcommand\shline{\noalign{\global\savewidth\arrayrulewidth
  \global\arrayrulewidth 1pt}\hline\noalign{\global\arrayrulewidth\savewidth}}
\newcommand{\tablestyle}[2]{\setlength{\tabcolsep}{#1}\renewcommand{\arraystretch}{#2}\centering\footnotesize}
\usepackage[breaklinks=true,bookmarks=false]{hyperref}

\iccvfinalcopy 

\def\iccvPaperID{10188} 

\ificcvfinal\fi

\begin{document}

\title{Exploring Temporal Coherence for More General Video Face Forgery Detection}

\author{Yinglin Zheng{\small $~^{1}$} ~Jianmin Bao{\small $~^{2}$}, ~Dong Chen{\small $~^{2}$}, ~Ming Zeng{\small $~^{1}$}\thanks{Corresponding author.}, ~Fang Wen{\small $~^{2}$} \\
\normalsize
$^{1}$ School of Informatics, Xiamen University \\
\normalsize
$^{2}$ Microsoft Research Asia\\

\normalsize
{\tt\small \{zhengyinglin@stu., zengming@\}xmu.edu.cn, \{jianbao, doch, fangwen\}@microsoft.com}
}


\maketitle
\ificcvfinal\thispagestyle{empty}\fi

\renewcommand{\thefootnote}{\fnsymbol{footnote}} 

\begin{abstract}
Although current face manipulation techniques achieve impressive performance regarding quality and controllability, they are struggling to generate temporal coherent face videos. 
In this work, we explore to take full advantage of the temporal coherence for video face forgery detection.
To achieve this, we propose a novel end-to-end framework, which consists of two major stages. The first stage is a fully temporal convolution network (FTCN). The key insight of FTCN is to reduce the spatial convolution kernel size to 1, while maintaining the temporal convolution kernel size unchanged. We surprisingly find this special design can benefit the model for extracting the temporal features as well as improve the generalization capability. The second stage is a Temporal Transformer network, which aims to explore the long-term temporal coherence.
The proposed framework is general and flexible, which can be directly trained from scratch without any pre-training models or external datasets.
Extensive experiments show that our framework outperforms existing methods and remains effective when applied to detect new sorts of face forgery videos.

\end{abstract}

\vspace{-4mm}
\section{Introduction}
\vspace{-2mm}

With the development of deep generative models, especially Generative Adversarial Networks (GANs)~\cite{korshunova2017fast,nirkin2019fsgan,bao2018towards,natsume2018rsgan,li2020advancing}. Current face manipulation techniques~\cite{li2020advancing, thies2020neural, thies2019deferred, thies2016face2face, jiang2020deeperforensics, vougioukas2019realistic, vougioukas2019end} are capable of manipulating the attributes or even the identity of face images. These forged images are even difficult to distinguish by humans, and thus may be abused for spreading political propaganda, damaging our trust in online media. Therefore, detecting face forgery is of paramount importance.

Most previous methods~\cite{zhou2017two,zhou2018learning,rossler2019faceforensics++,sabir2019recurrent} are trained for known face manipulation techniques. But they experience a dramatic drop in performance when the manipulation methods are unseen.
Some recent works~\cite{li2018exposing,xuan2019generalization,du2019towards,zhao2020learning,ganiyusufoglu2020spatio,qian2020thinking,masi2020two,chai2020makes} have noticed this problem and attempted to boost the generalization. However, these methods are vulnerable to common perturbations such as image or video compression, noise, and so on. They still show limited generalization capability. A particularly effective work is Face X-ray~\cite{li2020face}, which proposes to detect blending artifacts instead of generative artifacts. However, the blending artifacts are typically low-level information that is susceptible to post-processing operations. More recent work LipForensics~\cite{haliassos2020lips} proposes to detect unnatural mouth motion using spatio-temporal neural networks. But they only pay attention to the mouth which may ignore the artifacts in the other region of the face.


\begin{figure} 
    \centering 
    \includegraphics[width=\linewidth]{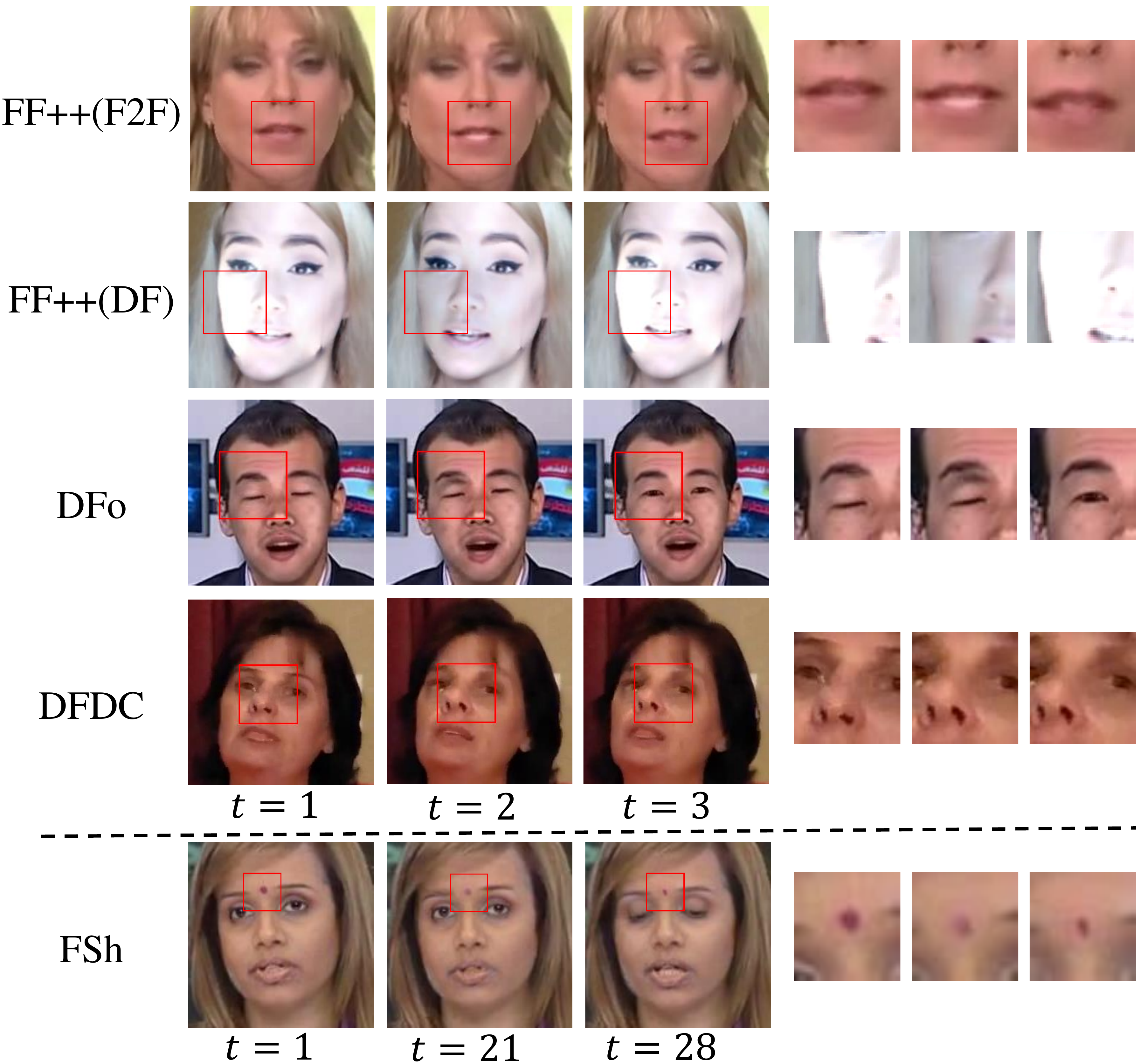}
    \caption{Temporal incoherence in existing datasets: FaceForensic++(FF++)~\cite{rossler2019faceforensics++}, DeeperForensics(DFo)~\cite{jiang2020deeperforensics}, Deepfake Detection Challenge Preview(DFDC)~\cite{dolhansky2020deepfake}, and FaceShifter(FSh)~\cite{li2020advancing}. In the top 4 rows, we show four temporal incoherence that happened between the neighborhood frames. In the last row, we show temporal incoherence happened between long-range frames.}
    \label{fig:temporal_incoherence_examples}
    \vspace{-2.0mm}
\end{figure}

In this paper, we propose to leverage temporal coherence for more general face forgery detection. We observe that most face video forgeries are generated in a frame-by-frame manner. Since each altered face is generated independently, it inevitably leads to obvious flickering and discontinuity of the face area (see Figure~\ref{fig:temporal_incoherence_examples}). So we can leverage the temporal incoherence for more general and robust video face forgery detection. Previous works try to leverage spatio-temporal convolution network~\cite{ganiyusufoglu2020spatio} or the recurrent neural network~\cite{amerini2020exploiting,masi2020two} to learn temporal incoherence. However, we find that they all failed to learn the general temporal incoherence.

After careful investigation, we find that forged face videos mainly contain two types of artifacts, one is spatially related (\eg blending boundary, checkboard, blur artifacts), the other is the temporal incoherence. Normally, the spatially related artifacts are more significant than the temporal incoherence. Without any specific design, current video face forgery detection methods~\cite{ganiyusufoglu2020spatio,amerini2020exploiting,masi2020two} may rely more on spatial-related artifacts instead of the temporal incoherence for classification. 

To encourage the spatio-temporal convolution network to learn the temporal incoherence, we redesign the convolution operator and propose a fully temporal convolution network (FTCN). The key idea is to restrict the network's capability for handling spatial-related artifacts. So we set the kernel size of all spatial (height and width) dimensions to 1 and keep the original kernel size of the time dimension in the 3D convolution operator. Due to the extremely low field for spatial dimension, the network learns to classify by temporal-related artifacts and hardly applies the spatial artifacts for detection. Also, we notice that even if the convolution operator is only time-dependent, its capability is sufficient to distinguish between real or fake.

Moreover, we find some discontinuity may happen in frames that are not in the neighborhood, for example, the wrinkles or moles of a face may gradually appear or disappear. To handle this issue, we propose to leverage Transformer~\cite{vaswani2017attention} for capturing long-range dependencies along the time dimension. We add a light-weight Temporal Transformer after the proposed FTCN. The FTCN and the Temporal Transformer are trained end-to-end as the whole framework for general video face forgery detection.


Our approach is general and flexible. It can achieve impressive results without any pre-training knowledge or hand-crafted datasets. In contrast, previous work LipForensics~\cite{haliassos2020lips} relies heavily on pre-training and Face X-ray relies on hand-crafted dataset. More importantly, without any manual annotations, our method can locate and visualize the temporal incoherence in the face forgery videos.


We conduct extensive experiments to compare its performance with the state-of-the-art in various challenging scenarios. We find that our method significantly outperforms
previous methods in terms of generalization capability to unseen forgeries, and robust to various perturbations on videos. Furthermore, we perform ablation studies to validate the design choices of our framework.

Our contributions are summarized as follows:


\begin{itemize}
    \vspace{-3mm}
    \item We explore to take full advantage of temporal coherence for face forgery detection and propose a framework that combined fully temporal convolution network (FTCN) and Temporal Transformer to explicitly detect temporal incoherence.
    
    \vspace{-2mm}
    \item Equipped with our detector, we can locate and visualize the temporal incoherence part of the face forgeries.
    
    \vspace{-2mm}
    \item Extensive experiments on various datasets demonstrate the superiority of our proposed methods with respect to generalization capability to unseen forgeries.
\end{itemize}

\vspace{-4mm}
\section{Related Work}
\vspace{-2mm}
Due to the emergence of high-fidelity face manipulation techniques, the detection of face forgery becomes an increasingly important research area. We will briefly introduce previous work on face forgery detection in this section.


\noindent \textbf{Image Face Forgery Detection.} Early studies put more emphasis on the spatial artifacts on the generated images, so they leverage the capability of convolution neural networks apply deep CNN models~\cite{afchar2018mesonet, bayar2016deep, hsu2018learning, rossler2019faceforensics++} to train a binary classifier to distinguish real or fake. Meanwhile, a significant amount of works explores low-level image statistics(\eg frequency, color)~\cite{qian2020thinking, li2018detection, durall2020watch, zhang2019detecting, frank2020leveraging} or high-level semantics(\eg identities)~\cite{nirkin2020deepfake} of face images for forgery detection. Some recent studies~\cite{bappy2019hybrid, agarwal2019protecting, salloum2018image, dang2020detection, Islam_2020_CVPR, Wu_2019_CVPR} aim to locate the visual artifacts in the forged images and make predictions based on the location results. 

\noindent \textbf{Video Face Forgery Detection.} More recently, a great number of works start to take the temporal dimension into consideration and conducting face forgery detection at the video level. Li \emph{et al}.~\cite{li2018ictu} introduce leveraging eye blinking for detecting generated fake face videos. Amerini \emph{et al}.~\cite{amerini2019deepfake} suggests using the optical flow between video frames. Mittal \emph{et al}.~\cite{mittal2020emotions} use the effective cues between audio and visual appearance for detecting fake videos. In contrast to these methods, our methods put more emphasis on general temporal incoherence. Unlike previous works, they either directly apply 3D convolution network ~\cite{de2020deepfake,ganiyusufoglu2020spatio} on video or detect a specific kind of incoherence, such as irregular eye blinking~\cite{li2018ictu}, lip motions~\cite{haliassos2020lips} or emotion~\cite{mittal2020emotions}. In this work, we seek to detect general temporal incoherence, which can be any inconsistent region along time dimension.


\begin{figure*}[ht] 
    \centering 
    \includegraphics[width=0.90\linewidth]{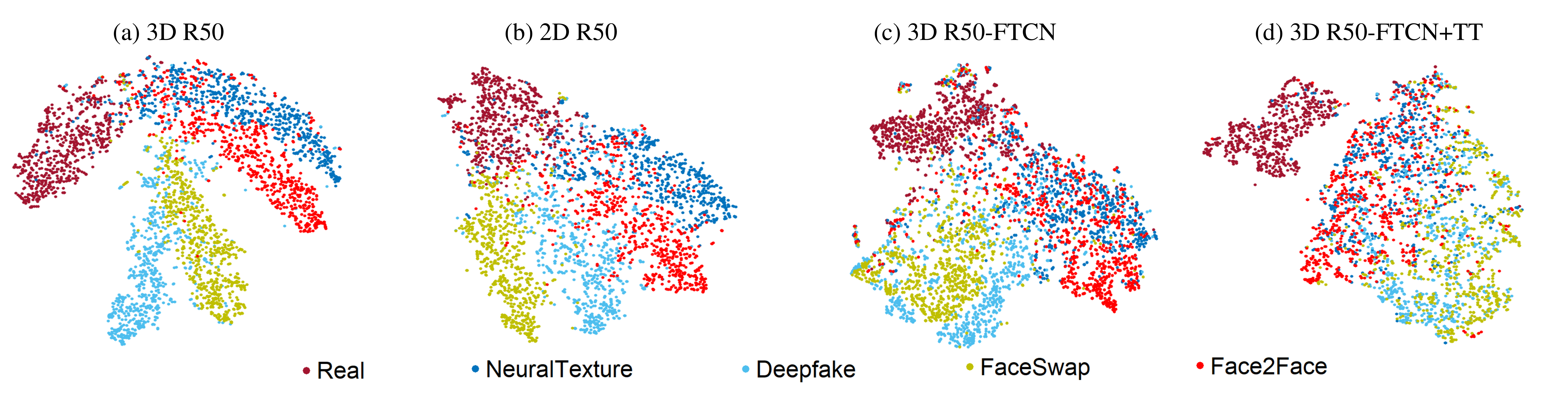}
    \caption{The t-SNE visualizations of features extracted from the last feature of different models on FF++ test set. Each dot represents the feature of a video clip. The t-SNE is computed with 40 for perplexity, 30 for PCA at 1000 iterations.}
    \label{fig:tsne}
    \vspace{-3mm}
\end{figure*}
\noindent \textbf{Generalization to Unseen Manipulations.} With the evolution of novel face manipulation techniques, many current detectors~\cite{rossler2019faceforensics++, afchar2018mesonet} can experience a significant performance drop. Many works have noted this problem and presented some solutions for improving the generalization capability of detectors.  FWA~\cite{li2018exposing} explores to detect the resolution differences between altered face and background for general deepfake detection. LAE~\cite{du2019towards} and Multi-task~\cite{nguyen2019multi} propose to learn a segmentation mask for the manipulated area in order to get a general detector. PatchForensics~\cite{chai2020makes} suggests that the patch-based classifier can improve the performance for generalized face forgery detection.
Face X-ray~\cite{li2020face} aims to improve the generalization capability by detecting the blending boundary artifacts instead of the artifacts from generative models. It can achieve impressive results in terms of generalization capability, but it is susceptible to many common perturbations. More recent LipForensics~\cite{haliassos2020lips} indicates that many current deepfake techniques may suffer from unnatural mouth motions, so they apply spatial-temporal networks for the detection. However, they ignore the other regions in the face region, which may damage the performance.

\vspace{-3mm}
\section{Methods}
\vspace{-1mm}
\subsection{Motivation}
\vspace{-1mm}
Currently, most face manipulation methods~\cite{li2020advancing, thies2020neural, thies2019deferred, thies2016face2face, vougioukas2019realistic, vougioukas2019end} are specifically crafted at the image level. To generate a fake video, current techniques need to apply their methods for each frame independently. However, subtle changes in the appearance (\eg noise, lighting, motion) often result in temporal incoherent results (\eg flickering and discontinuous results as shown in Figure~\ref{fig:temporal_incoherence_examples}). Some prior face manipulation techniques~\cite{jiang2020deeperforensics, Celeb_DF_cvpr20} have noticed this problem and apply post-processing tools to tackle this issue, but the generated videos still suffer from the problem of temporal incoherence to some extent.
Therefore, how to detect temporal incoherence is worthy of more careful study.

Detecting the temporal incoherence is challenging since we do not have the location annotation of the incoherence in the video. A naive idea is to adopt the spatio-temporal convolution networks~\cite{ganiyusufoglu2020spatio, de2020deepfake} and expect the model to learn to distinguish real or fake by temporal incoherence.
However, we find that forged face videos mainly contain two types of artifacts, one is the spatially related (\eg blending boundary, checkboard, blur artifacts), the other is the temporal incoherence. Normally, the spatially related artifacts are more significant than the temporal incoherence. 
Without any specific design, the spatio-temporal convolution network distinguishes real or fake using spatial artifacts instead of temporal incoherence. 

So the problem becomes how to encourage the spatio-temporal convolution network to learn the temporal incoherence. We take a fundamentally different approach, we propose a fully temporal convolution network. Concretely, we keep all the temporal-related convolution kernel size as the original but set all the spatial-related convolution kernel size to $1$. We find this restriction can encourage the network to learn the temporal incoherence. To prove that, we take ResNet-50(R50)~\cite{he2016deep} as backbone and compare three types of classifiers:

1. A 3D R50~\cite{hara2017learning} network structure, which employs spatio-temporal convolutions.

2. A 2D R50 network structure, which uses 2D convolutions.

3. The proposed 3D R50-FTCN. We use 3D R50 as the backbone and set all the spatial-related convolution kernel size to 1 and keep the temporal-related kernel size.

To ensure fair comparison, all classifiers use the same training set FF++~\cite{rossler2019faceforensics++} and the same training and inference settings. We show the t-SNE visualization of features extracted from different classifiers on FF++ test set in Figure~\ref{fig:tsne}. We have the following observations:
Although all classifiers can distinguish between real and fake data, the distribution of fake data is completely different. Both the 3D R50 and 2D R50 will separate fake data generated by different face manipulation methods, even if we treat all fake data as one class in the training stage. It clearly shows that the features they extract contain the unique artifacts of each face manipulation algorithm. This would affect their generalization ability. On the contrary, the fake data of the 3D R50-FTCN classifier are more mixed together. It proves that the temporal network learns to classify by more general temporal incoherence.

On the other hand, some temporal incoherence exists in the \emph{long-range} of video frames. However, prior studies~\cite{wang2018non, hu2018relation} indicate that temporal convolution struggles in dealing with the \emph{long-range} dependencies. To tackle this issue, we add a Temporal Transformer(TT)~\cite{dosovitskiy2020image} after the FTCN to detect the long-range temporal incoherence. The Temporal Transformer takes sequences of temporal features extracted by the FTCN as input and apply the class token to make the predictions. We also show the feature distribution of this classifier in Figure~\ref{fig:tsne} (denoted as 3D R50-FTCN+TT). Using Transformer can further separate real and fake data, and can further gather the features of different face manipulation algorithms.


\vspace{-1mm}
\subsection{Overall Framework}
\vspace{-2mm}

In this section, we introduce the details of the proposed Fully Temporal Convolution network and Temporal Transformer. These two parts are trained end-to-end for video face forgery detection. Overall, given a suspect video $\mbf{V}$, the first stage is the Fully Temporal Convolution Network (FTCN) that deals with local temporal flickering and inconsistency. It extracts temporal feature $\mbf{F}=\mathbf{FTCN}(\mbf{V})$. The second stage is the Temporal Transformer that aims to further model the long-term incoherence between each time slice of $\mbf{F}$. Finally, an MLP head is used to do the final prediction.

\newcommand{\blockb}[3]{\multirow{3}{*}{\(\left[\begin{array}{c}\text{1\x1\x1, #2}\\[-.1em] \text{\textbf{3\x1\x1}, #2}\\[-.1em] \text{1\x1\x1, #1}\end{array}\right]\)\x#3}}
\begin{table}[t]
\footnotesize
\centering
\resizebox{0.89\columnwidth}{!}{
\tablestyle{6pt}{1.08}
\begin{tabular}{c|c|c}
\multicolumn{2}{c|}{layer} & output size \\
\shline
conv$_1$ & \multicolumn{1}{c|}{\textbf{5\x1\x1}, 64, stride 1, 1, 1} & 64\x32\x224\x224 \\
\hline
pool$_1$  & \multicolumn{1}{c|}{1\x5\x5 max, stride 1, 4, 4} & 256\x32\x56\x56 \\
\hline
\multirow{3}{*}{res$_2$} & \blockb{256}{64}{3} & \multirow{3}{*}{256\x32\x56\x56} \\
  &  & \\
  &  & \\
\hline
pool$_2$  & \multicolumn{1}{c|}{2\x1\x1 max, stride 2, 1, 1} & 256\x16\x56\x56 \\
\hline
\multirow{3}{*}{res$_3$} & \blockb{512}{128}{4} & \multirow{3}{*}{512\x16\x28\x28} \\
  &  & \\
  &  & \\
\hline
\multirow{3}{*}{res$_4$} & \blockb{1024}{256}{6} & \multirow{3}{*}{1024\x16\x14\x14}  \\
  &  & \\
  &  & \\
\hline
\multirow{3}{*}{res$_5$} & \blockb{2048}{512}{3} & \multirow{3}{*}{2048\x16\x7\x7} \\
  &  & \\
  &  & \\
\hline
\multicolumn{2}{c|}{spatial-related average pool} & 2048\x16\x1\x1  \\
\end{tabular}}
\vspace{.5mm}
\caption{Our 3D R50-FTCN model for video face forgery detection. Compared with the original 3D ResNet-50 model~\cite{carreira2017quo}, we follow the rules described in Section~\ref{sec:ftcn} to obtain this structure. The dimension of 3D filter kernels are in $K_t$\x$K_h$\x$K_w$, The dimension of output maps are in $C$\x$N$\x$H$\x$W$. The input is 32\x224\x224. Residual blocks are shown in brackets. 
}
\vspace{-4mm}
\label{tab:FTCN}
\end{table}

\vspace{-2mm}
\subsubsection{Fully Temporal Convolution Network}
\vspace{-2mm}
\label{sec:ftcn}

3D CNNs~\cite{carreira2017quo, tran2018closer} are widely used on video-related tasks. Traditional 3D CNN models compute both spatial-temporal correlations via convolution with $K_t$\x$K_h$\x$K_w$ kernel, where $K_t$\x$K_h$\x$K_w$ are designed to be larger than 1 for most layers. These convolution layers are applied repeatedly, propagating signals progressively through the spatial and temporal dimensions. However, we find such spatial-temporal coupled kernels weaken the model's ability to capture purely temporal information. To encourage the spatio-temporal convolution network to learn the temporal incoherence, we redesign the convolution operator and propose a fully temporal convolution network (FTCN). The key idea is to restrict the capability of the network for handling spatial-related artifacts. So we set all the spatial sizes of the convolution kernel to $1$. As we notice that some convolution layers may involve strides more than 1. In this situation, many locations may be ignored in the input features, thus we also design a rule to handle this. Suppose a 3D convolution is represented as 3DConv($K_t$, $K_h$, $K_w$, $S_t$, $S_h$, $S_w$), where $K_t$, $K_h$, $K_w$ are the kernel size for time, height, width dimension, $S_t$, $S_h$, $S_w$ are the stride for time, height, width dimension. We replace it with 3DConv($K_t$, 1, 1, 1, 1, 1) and add a max-pooling after the convolution operator if $S_h$ or $S_w > 1$.


Take the popular 3D R50 structure~\cite{carreira2017quo} as an example, the resulting 3D R50-FTCN model is shown in Table~\ref{tab:FTCN}. The input video clip has 32 frames each with 224\x224 pixels. Also, we change the last global average pooling to spatial-related average pooling to keep the feature along time dimension unchanged. All the 3D convolutions in Table~\ref{tab:FTCN} share a similar kernel shape with like $K_t$\x1\x1, which can be regarded as 1D temporal convolutional filters for temporal dimension. Such networks can be regarded as fully temporal convolution networks (FTCN), which mainly learn the discriminative features along the temporal dimension. 



\begin{figure} 
	\centering 
	\includegraphics[width=\linewidth]{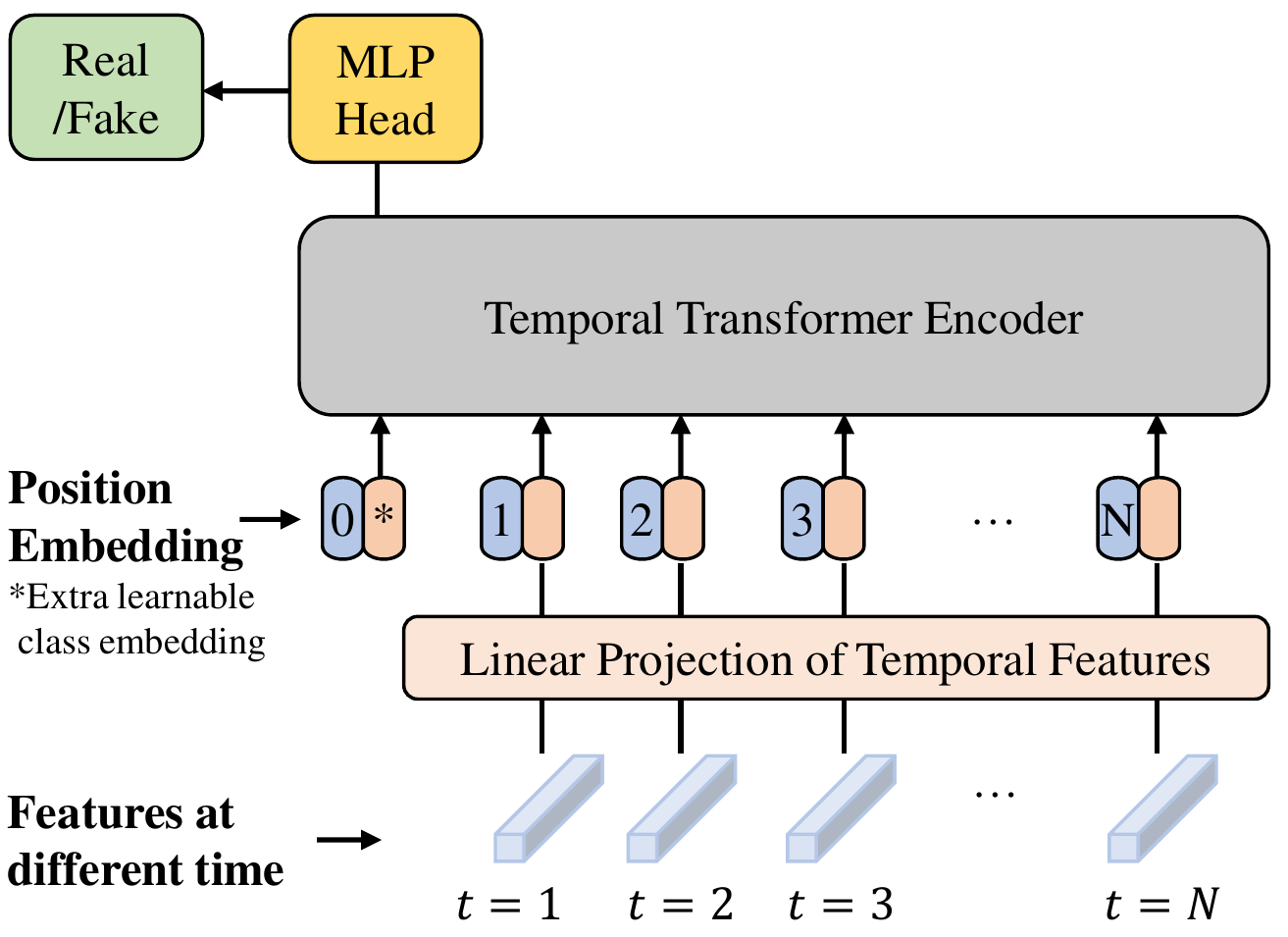}
	\caption{The Temporal Transformer for our video face forgery detection framework. We split the feature extracted from FTCN along the time dimension, and feed the resulting sequences of features to a standard Transformer encoder. We apply an extra learnable ``classification token" to the sequence to learn the final discriminative feature and add an MLP head to distinguish real or fake.}
	\label{fig:Transformer}
	\vspace{-3mm}
\end{figure}

\vspace{-2mm}
\subsubsection{Temporal Transformer}
\vspace{-2mm}

The Temporal Transformer aims to learn the \emph{long-range} discrepancies along the time dimension. With the FTCN, we obtain the temporal features $\mbf{F}\in {R}^{C\times N\times H\times W}$~($C$=2048, $N$=16, $H$=1, $W$=1). The temporal feature $\mbf{F}$ can be represented as a sequence of features $\mbf{F_t}\in {R}^{C}, t \in \{1,2,...,N\}$, which is a 1D sequence of token embeddings in standard Transformer~\cite{vaswani2017attention}. The  $
N$ is input sequence length, $C$ is the feature dimension of sequence. The overview of our Temporal Transformer is depicted in Figure~\ref{fig:Transformer}.

Similar to the settings in ViT~\cite{dosovitskiy2020image}, we apply a trainable liner projection $\mbf{W}$ to map the feature dimension from $C$ to $D$. To enable the classification in Temporal Transformer, we add a learnable embedding to the sequence of embedded features ($\mbf{z_0^0} = \mbf{F_{class}}$), which serves as the representative features learned on the input sequences. Following the settings in ViT\cite{dosovitskiy2020image}, we also include learnable 1D position embeddings to retain positional information. Suppose the position embedding is $\mbf{E_{pos}}$. So the input sequence $z_0$ for the Temporal Transformer can be defined as:
\begin{equation}
    \mbf{z}_0 = [ \mbf{F}_\text{class}, \mbf{W} \mbf{F}_1 , \mbf{W} \mbf{F}_2 , \cdots, \mbf{W} \mbf{F}_{N}]^T + \mbf{E}_{pos},
    \label{eqn:embedding}
\end{equation}
where $\mbf{F}_t$ is the $t$-th time slice in feature $\mbf{F}$, $\mbf{W} \in \mathbb{R}^{D \times C}, \mbf{E}_{pos}  \in \mathbb{R}^{(N + 1) \times D}$.

The Temporal Transformer mainly consists of $L$ standard Transformer Encoder blocks~\cite{vaswani2017attention}, each standard Transformer encoder block consists of a multi-head self-attention(MSA)~\cite{vaswani2017attention} block and an MLP block. We also apply the commonly-used LayerNorm(LN) before each block. Another important structure is the residual connections~\cite{he2016deep}, which is applied to each block. The activation function we used for the Temporal Transformer is GELU. So the features get for the $l$-th layer can be defined as:
\begin{align}
  \mbf{z^\prime}_\ell &= \op{MSA}(\op{LN}(\mbf{z}_{\ell-1})) + \mbf{z}_{\ell-1}, && \ell=1\ldots L  \\
    \mbf{z}_\ell &= \op{MLP}(\op{LN}(\mbf{z^\prime}_{\ell})) + \mbf{z^\prime}_{\ell}, && \ell=1\ldots L  
\end{align}
So based on the class-token output $\op{LN}(\mbf{z}_L^0)$ of the last encoder. We can apply an MLP head for the final fake probability:
\begin{equation}
    y = \op{MLP}(\op{LN}(\mbf{z}_L^0)).
\end{equation}
In our experiments, the binary cross-entropy loss is applied on the final prediction $y$.


\vspace{-3mm}
\section{Experiments}
\vspace{-1mm}
\subsection{Experimental Settings}
\vspace{-1mm}
\noindent\textbf{Training Datasets.} We adopt the most commonly used benchmark dataset FaceForensics++(FF++)\cite{roessler2019faceforensicspp} for training. It contains 1000 original videos and 4000 fake videos. The fake videos are manipulated by four methods: Face2Face(F2F)~\cite{thies2016face2face}, FaceSwap(FS)~\cite{faceswap}, NeuralTexture(NT)~\cite{thies2019deferred}, and Deepfake(DF)~\cite{deepfakes}. We trained on the high-quality (HQ) subset of the FF++, which is the light compression version.

\noindent\textbf{Testing Datasets.} To evaluate the generalization capability of our framework, we test our model on the following datasets: 1) FF++ that contains four types of manipulation as mentioned above; 2) FaceShifter\cite{li2020advancing}(FSh) and 3) DeeperForensics\cite{jiang2020deeperforensics}(DFo) employ the real videos from FF++ for high-fidelity face swapping; 4) DeepFake Detection Challenge Preview dataset\cite{dolhansky2020deepfake}(DFDC), where each original video is filmed in the challenging environment; and 5) Celeb-DF-v2\cite{Celeb_DF_cvpr20}(CDF) is a new DeepFake dataset including 518 videos from different sources.

\noindent\textbf{Evaluation Metrics.} Following the evaluation metrics in previous works~\cite{li2020face, haliassos2020lips}, we report the area under the receiver operating characteristic curve (AUC). Since most previous works are image-based, following the setting in LipForensics~\cite{haliassos2020lips}, we report video-level AUC for a fair comparison. For the image-based method, we average the model predictions for each frame across the entire video. Therefore, all models utilize the same number of frames for classification.

\noindent\textbf{Implementation.} We take the 3D R50 as the basic structure of our proposed FTCN. The Temporal Transformer we use is one layer of standard Transformer Encoder~\cite{vaswani2017attention}, whose self-attention heads, hidden size, and MLP size are set to 12, 1024, 2048, respectively. For the training setting, we use a batch size of 32 and SGD optimizer with momentum, and the weight decay is set as 1e-4. We apply a warm-up strategy for the training of our methods. Concretely, the learning rate first increases from 0.01 to 0.1 in the first 10 epochs and then cosinely decayed to 0 for the last 90 epochs. For more details, please refer to the supplementary material.

\noindent\textbf{Baselines.} We mainly compare our methods with various state-of-the-art methods. These methods are mainly about boosting the generalization capability as well as some popular baselines.
1) \textbf{Xception}~\cite{rossler2019faceforensics++} explores the performance of face manipulation detection by the popular Xception~\cite{chollet2017xception} model.
2) \textbf{CNN-aug}~\cite{wang2020cnn} find that current CNN-generated images can be easily detected by a CNN model.
3) \textbf{PatchForensics}~\cite{chai2020makes} suggests that the patch-based classifier can achieve impressive results for face forgery detection.
4) \textbf{Face X-ray}~\cite{li2020face} aims to improve the generalization capability by detecting the blending boundary artifacts instead of the artifacts from generative models.
5) \textbf{CNN-GRU}~\cite{sabir2019recurrent} introduces GRU~\cite{cho2014learning} into CNN model to model temporal coherence.
6) \textbf{Multi-task}~\cite{nguyen2019multi} applies an autoencoder-like architecture for deepfake detection.
7) \textbf{FWA}~\cite{li2018exposing} explores to detect the resolution differences between altered face and background for improving generalization capability.
8) \textbf{Two-branch}~\cite{masi2020two} presents multi-task learning on FaceForensics++ dataset.
9) \textbf{LipForensics}~\cite{haliassos2020lips} is a recent work that studies the irregular mouth motions for general and robust face forgery detection.

\vspace{-2mm}
\subsection{Generalization to unseen manipulations}
\vspace{-2mm}
The differences between face forgery datasets mainly lie in the variations of source videos and face manipulation methods. To evaluate the cross-manipulations generalization capability of different face forgery detectors and prevent the possible bias introduced by different source videos, we conduct experiments on FF++, as it provides fake videos created by multiple face forgery methods for the same source videos. Following the setting in \cite{haliassos2020lips}, we evaluate face forgery detectors with the leave-one-out strategy. To be concrete, as there are four types of fake video in FF++, each type is used once as a test set while the remaining three types form the training set. Both training and testing are conducted on the HQ version of FF++ dataset.

Table~\ref{tab:cross-method} shows that our method achieves excellent generalization(99.7\%) to novel forgeries, surpassing on average most approaches by large margins. Although four types of manipulations(Deepfake, FaceSwap, Face2Face, NeuralTexture) in FF++ use different methods and focus on different tasks, our framework can learn a generalized discriminative feature on three of the manipulations and generalized to the remaining one. Our framework outperforms recent state-of-the-art methods Face X-ray~\cite{li2020face} and LipForensics~\cite{haliassos2020lips} by 4.8\% and 2.6\% in terms of AUC, respectively. We also present the number of parameters of Face X-ray and LipForensics on Table \ref{table:compare_params}, our method achieves the highest performance with the minimal number of parameters, without any pre-training or external training data, which further demonstrate the superiority of our framework.




\begin{table}[]
\footnotesize
\centering
\resizebox{1.02\columnwidth}{!}{
\tablestyle{7pt}{1.08}
\begin{tabular}{lccccc}

\hline
\multirow{2}{*}{Method} & \multicolumn{4}{c}{Train on remaining three} &      \\ \cline{2-5}
                        & DF        & FS        & F2F       & NT       & \textbf{Avg}  \\ \hline
Xception ~\cite{rossler2019faceforensics++}              & 93.9      & 51.2      & 86.8      & 79.7     & 77.9 \\
CNN-aug   ~\cite{wang2020cnn}              & 87.5      & 56.3      & 80.1      & 67.8     & 72.9 \\
PatchForensics\cite{chai2020makes}             & 94.0      & 60.5      & 87.3      & 84.8     & 81.7 \\
CNN-GRU  ~\cite{sabir2019recurrent}              & 97.6      & 47.6      & 85.8      & 86.6     & 79.4 \\
Face X-ray\cite{li2020face}             & 99.5      & 93.2      & 94.5      & 92.5     & 94.9 \\
\hline
LipForensics-Scratch\cite{haliassos2020lips}& 93.0      & 56.7      & 98.8      & 98.3     & 86.7 \\
LipForensics\cite{haliassos2020lips}            & 99.7      & 90.1      & \textbf{99.7}      & 99.1     & 97.1 \\ \hline
ours                    &   \textbf{99.9}        &  \textbf{99.9}         & \textbf{99.7}          & \textbf{99.2}         &    \textbf{99.7}  \\ \hline
\end{tabular}}
\vspace{0.5mm}
\caption{\textbf{Generalization to unseen manipulations.} We report the video-level AUC(\%) on the FF++ dataset, which consists of four manipulation methods(DF, FS, F2F, NT). We train on three methods and test on the other one method. The results of other methods are from ~\cite{haliassos2020lips}.}
\label{tab:cross-method} 
\vspace{-3mm}
\end{table}

\begin{table}[]
\centering
\resizebox{1.0\columnwidth}{!}{
\tablestyle{7pt}{1.08}
\begin{tabular}{lcccc}
\hline
\multicolumn{1}{c}{Model}                                     & \multicolumn{1}{l}{\#params} & \multicolumn{1}{l}{Pre-train} & \multicolumn{1}{l}{Extra data} & \multicolumn{1}{l}{Avg} \\ \hline
Face X-ray\cite{li2020face}                  & 65.8M                        & N                            & Y                              & 94.9                    \\
LipForensics-Scrtach\cite{haliassos2020lips} & 36.0M                        & N                            & N                              & 86.7                    \\
LipForensics\cite{haliassos2020lips}         & 36.0M                        & Y                            & N                              & 97.1                    \\ \hline
ours                                                          & \textbf{26.6M}                        & N                            & N                              & \textbf{99.7}                    \\ \hline
\end{tabular}}
\vspace{0.5mm}
\caption{\textbf{Comparison with state-of-the-art methods.} We report the number of parameters, pre-training, extra data usage, and the average video level AUC(\%)  on four unseen manipulation methods(DF, FS, F2F, NT), all the models are trained on the remaining three methods in FF++.}
\label{table:compare_params}
\vspace{-3mm}
\end{table}

\vspace{-2mm}
\subsection{Generalization to unseen datasets}
\vspace{-2mm}
In a real-world scenario, suspicious videos are likely to be created by unseen methods from unseen source videos, thus the cross-dataset generalization would be crucial. To evaluate the cross-dataset generalization capability, we trained the face forgery detectors on all the four types of fake data in FF++, and perform the evaluation on four unseen datasets, including Celeb-DF-v2(CDF)~\cite{Celeb_DF_cvpr20}, DFDC~\cite{dolhansky2020deepfake}, FaceShifter~\cite{li2020advancing} and DeeperFoensics~\cite{jiang2020deeperforensics}. As shown in Table \ref{tab:cross-dataset}, our model achieves the best performance on every dataset, with especially strong results on FaceShifter and DeeperForensics. On CDF and DFDC dataset, all the methods obtain relatively low scores, one possible explanation is the scenario gaps between different datasets.

\begin{table}[]
\footnotesize
\centering
\resizebox{1.0\columnwidth}{!}{
\tablestyle{7pt}{1.08}
\begin{tabular}{lccccc}
\hline
Method       & CDF  & DFDC                  & FSh                   & DFo                   & \textbf{Avg}          \\ \hline
Xception~\cite{rossler2019faceforensics++}     & 73.7 & 70.9                  & 72.0                  & 84.5                  & 75.3                  \\
CNN-aug~\cite{wang2020cnn}      & 75.6 & 72.1                  & 65.7                  & 74.4                  & 72.0                  \\
PatchForensics~\cite{chai2020makes}  & 69.6 & 65.6                  & 57.8                  & 81.8                  & 68.7                  \\
CNN-GRU~\cite{sabir2019recurrent}      & 69.8 & 68.9                  & 80.8                  & 74.1                  & 73.4                  \\
Multi-task~\cite{nguyen2019multi}   & 75.7 & 68.1                  & 66.0                  & 77.7                  & 71.9                  \\
 FWA~\cite{li2018exposing}      & 69.5 & 67.3                  & 65.5                  & 50.2                  & 63.1                  \\
Two-branch~\cite{masi2020two}   & 76.7 &   ---                    & ---                      & ---                      &    ---                   \\
Face X-ray~\cite{li2020face}   & 79.5 & 65.5                  & 92.8                  & 86.8                  & 81.2                  \\
LipForensics~\cite{haliassos2020lips} & 82.4 & 73.5                  & 97.1                  & 97.6                  & 87.7                  \\ \hline
ours         & \textbf{86.9} & \textbf{74.0}                  & \textbf{98.8}                  & \textbf{98.8}                  & \textbf{89.6}                     \\ \hline
\end{tabular}}
\vspace{0.5mm}
\caption{\textbf{Generalization to unseen datasets.} We report the video-level AUC(\%) on four unseen datasets: Celeb-DF-v2(CDF), Deepfake Detection Challenge Preview(DFDC), FaceShifter(FSh), and DeeperForensics(DFo). We train on FF++ and test on these unseen datasets. The results of other methods are from ~\cite{haliassos2020lips}.}
\label{tab:cross-dataset} 
\vspace{-3mm}
\end{table}

\vspace{-2mm}
\subsection{Robustness to unseen perturbations}
\begin{figure*}
\centering
	\includegraphics[scale=.51]{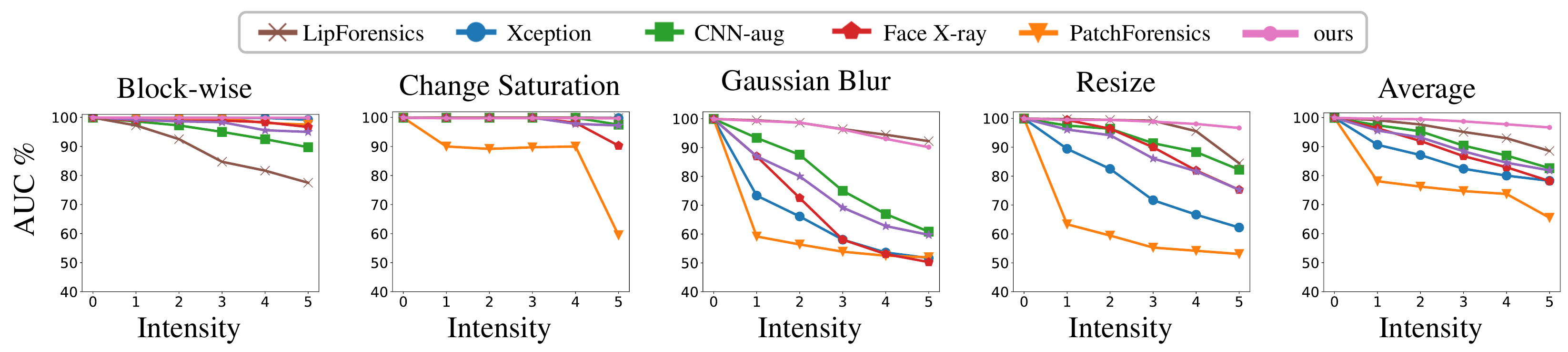}
\caption{\textbf{Robustness to unseen perturbations.} We report the video-level AUC(\%) of our methods under 5 different levels of four particular types of perturbations: Block-wise distortion, Change Saturation, Gaussian Blur, and Resize.}
\label{fig:rubostness}
\vspace{-3mm}
\end{figure*}

\vspace{-2mm}
For real-world scenarios, it is of great importance for face forgery detectors to be robust to unseen perturbations. We conduct experiments to verify the robustness of our methods.
Following \cite{jiang2020deeperforensics}, we consider four popular perturbations: 1) Block-wise distortion; 2) Change of color saturation;  3) Gaussian Blur; 4) Resize: downsample the image by a factor then upsample it to the original resolution. Each perturbation is divided into five intensity levels as \cite{jiang2020deeperforensics}. The results are reported in Figure~\ref{fig:rubostness}. On average, our methods achieve better robustness to unseen perturbations. It is worth mentioning that our method does not apply any pre-training knowledge during training, which is obviously helpful for robustness. 

\vspace{-2mm}
\subsection{Ablation study}
\vspace{-2mm}
We perform comprehensive studies on the FF++~\cite{rossler2019faceforensics++} dataset to validate our design of the overall framework. We start by verifying the design of 3D R50-FTCN.
\noindent \textbf{Why remove the spatial convolution?} To validate why we remove all the spatial convolution in the proposed 3D R50-FTCN, we construct multiple variants of 3D ResNet-50(3D R50), including the following models: 

1. \textbf{3D R50}: The original model of 3D R50, with spatio-temporal convolutions. 

2. \textbf{3D R50-Spatial}: Based on 3D R50, replace all the 3DConv($K_t$, $K_h$, $K_w$, 1, 1, 1) with 3DConv(1, $K_h$, $K_w$, 1, 1, 1). 


3. \textbf{3D R50-FTCN-FK3}: We replace the first 3D convolution layer of 3D R50-FTCN with 3DConv($5$, $3$, $3$, 1, 1, 1), which involves spatial-related convolution.

4. \textbf{3D R50-FTCN-FK5}: We replace the first convolution of 3D R50-FTCN with 3DConv($5$, $5$, $5$, 1, 1, 1), which involves more spatial-related convolution than \textbf{3D R50-FTCN-FK3}.

5. \textbf{3D R50-FTCN-Shuffle}: We use the same network structure as 3D R50-FTCN, but take a ``spatial-shuffled'' clip as input. The spatial-shuffle operation shuffles the pixel order in the clips along the spatial dimensions while keeping the pixel-wise temporal continuity~(all images in the clip share the same shuffle pattern), we show the shuffled clips in the Supplementary Material.

6. \textbf{3D R50-FTCN}: The proposed FTCN, its structure is described in Section \ref{sec:ftcn}, and shown in Table \ref{tab:FTCN}.

We apply these models to train a binary classifier and evaluate the performance under the setting of training only on F2F, and testing on all four methods of FF++~(F2F, FS, DF, NT). 
All the variants are trained with exactly the same setting except the model architecture as described above. For each model, we report the performance of the best checkpoint, which is selected based on the average AUC among four methods on the validation set. 

The results are shown in Table \ref{tab:whyspatial}. We can identify some important conclusions from the results: 1) Compare the results of 3D R50 and 3D R50-Spatial, involving temporal information into face forgery detection can benefit the performance of generalization. 2) For 3D R50, 3D R50-FTCN-FK5, 3D R50-FTCN-FK3, 3D R50-FTCN, the spatial-related convolution involves less and less, but the generalization capability gets better and better, so less spatial-related convolution leads to better results. 3) Even if we damage the spatial information by pixel shuffle, the 3D R50-FTCN can still achieve a reasonable result, this suggests that the 3D R50-FTCN mainly learns to distinguish by temporal-related information.

%

\begin{table}[]
\footnotesize
\centering
\resizebox{1.0\columnwidth}{!}{
\tablestyle{7pt}{1.08}
\begin{tabular}{llllll}
\hline
\multirow{2}{*}{Model} & \multicolumn{4}{c}{Train on F2F}                             &               \\ \cline{2-5}
                       & DF            & FS            & F2F          & NT            & Avg           \\ \hline
3D R50                 & 80.0          & 89.5          & \textbf{100} & 91.6          & 90.3          \\
3D R50-Spatial         & 77.9          & 53.6          & \textbf{100} & 79.1          & 77.7          \\
3D R50-FTCN-FK3        & 97.4          & 94.1          & \textbf{100} & 95.3          & 96.8          \\
3D R50-FTCN-FK5        & 94.2          & 93.0          & \textbf{100} & 93.2          & 95.1          \\
3D R50-FTCN-Shuffle    & 97.3          & 92.5          & \textbf{100} & 93.2          & 95.8          \\
3D R50-FTCN            & \textbf{98.0} & \textbf{95.9} & \textbf{100}          & \textbf{96.0} & \textbf{97.5} \\ \hline
\end{tabular}
}
\vspace{0.5mm}
\caption{\textbf{Ablation study of variants design of FTCN} Video-level AUC(\%) is reported on FF++ datasets.}
\label{tab:whyspatial}
\vspace{-4mm}
\end{table}

\vspace{1.0mm}
\noindent \textbf{Is limited model capability that benefits generalization capability?} For 3D R50, 3D R50-FTCN-FK5, 3D R50-FTCN-FK3, 3D R50-FTCN, the spatial-related convolution involves less and less, and the model capability becomes weaker and weaker. Thus it is natural to ask whether the generalization capability benefits from the limited model capability. We conduct experiments to verify this. We design several variants of our proposed 3D R50-FTCN and trained on the F2F in FF++:

1. \textbf{3D R50-SP}: 3D R50 with the same amount of parameters as 3D R50-FTCN, this model is created by reducing the number of channels in 3D R50.


2. \textbf{3D R50-FHCN}: Replace 3DConv($K_t$, $K_h$, $K_w$, $S_h$, $S_w$) with 3DConv(1, $K_h$, 1, 1, 1, 1) and add a MaxPool(1, $S_h$, $S_w$) if $S_h>1$ or $S_w > 1$. 

3. \textbf{3D R50-FWCN}: Replace 3DConv($K_t$, $K_h$, $K_w$, $S_h$, $S_w$) with 3DConv(1, 1, $K_w$, 1, 1, 1) and add a MaxPool(1, $S_h$, $S_w$) if $S_h>1$ or $S_w > 1$. 

The results are reported in Table~\ref{tab:similar_complexity}, 3D R50-SP and 3D R50-FTCN share a similar number of parameters, but 3D R50-FTCN presents better results. This validates that the performance gain is mainly from the fully temporal design. Moreover, 3D R50-FHCN and 3D R50-FWCN share exactly the same amount of parameters and computation cost with 3D R50-FTCN but present lower performance. This further indicates that temporal artifacts are more general as well as the effectiveness of our proposed 3D R50-FTCN.


\begin{table}[]
\footnotesize
\centering
\resizebox{1.0\columnwidth}{!}{
\tablestyle{7pt}{1.08}
\begin{tabular}{lccccc}
\hline
\multirow{2}{*}{Model} & \multicolumn{4}{c}{Train on F2F} &                          \\ \cline{2-5}
                       & DF     & FS     & F2F    & NT    & Avg                      \\ \hline
3D R50                 & 80.0   & 89.5   & \textbf{100}    & 91.6  & 90.3                     \\
3D R50-SP            & 86.2   & 85.3   & \textbf{100}    & 86.7  & \multicolumn{1}{l}{89.6} \\
3D R50-FHCN          & 76.1   & 49.5   & \textbf{100}    & 82.7  & 77.1                     \\
3D R50-FWCN           & 84.8   & 73.2   & 99.5   & 76.2  & 83.4                     \\ 
3D R50-FTCN            & \textbf{98.0} & \textbf{95.9} & \textbf{100}          & \textbf{96.0} & \textbf{97.5}                     \\ \hline
\end{tabular}}
\vspace{0.5mm}
\caption{\textbf{Ablation study of 3D R50 variants with different model capability.} Video-level AUC(\%) is reported.}
\label{tab:similar_complexity}
\vspace{-3mm}
\end{table}

\noindent \textbf{Influence of video clip size.} To find the optimal clip size, we trained 3D R50-FTCN with clip sizes of 8, 16, 32, 64. All the models are trained on F2F and test on all four methods in FF++. We change only the clip size and keep the unrelated hyper-parameters unchanged.

Table \ref{tab:clipsize} shows that as clip size increases, the performance boost. There is a tiny performance gain when clip size changes from 32 to 64, the possible reasons could be 1) There are not enough temporal convolution layers to capture such a long clip. 2) Video face alignment suffers from large clip size and large motion, as it would be hard to find a crop region that covers all the faces in the clip. As clip size growth brings performance boost along with more computation cost, a good trade-off would be clip size 32.

\begin{table}[]
\footnotesize
\centering
\resizebox{1.0\columnwidth}{!}{
\tablestyle{7pt}{1.08}
\begin{tabular}{clllll}
\hline
\multicolumn{1}{l}{\multirow{2}{*}{clip size}} & \multicolumn{4}{c}{Train on F2F} &      \\ \cline{2-5}
\multicolumn{1}{l}{}                           & DF     & FS     & F2F    & NT    & Avg  \\ \hline
8                                              & 79     & 86.2   & 99.8   & 85.6  & 87.7 \\
16                                             & 95.4   & 95.3   & \textbf{100}    & 94.8  & 96.4 \\
32                                             & 98.0   & 95.9   & \textbf{100}    & 96.0  & 97.5 \\
64                                             & \textbf{98.2}   & \textbf{96.6}   & \textbf{100}    & \textbf{96.7}  & \textbf{97.9} \\ \hline
\end{tabular}
}
\vspace{0.5mm}
\caption{\textbf{Ablation study of using different clip sizes for the training of FTCN.} Video-level AUC(\%) is reported.}
\label{tab:clipsize}
\vspace{-3mm}
\end{table}

\noindent \textbf{Effectiveness of Transformer.}
To validate the effectiveness of our Temporal Transformer, we perform an ablation study on the framework. We train three variants of our framework: 1) we train a model which only has the 3D R50 FTCN; 2) based on our framework, we change the number of encoder layers in Temporal Transformer to 2 (3D R50 FTCN+TT L\x 2); 3) based on our framework, we change the number of encoder layers in Temporal Transformer to 3 (3D R50 FTCN+TT L\x 3). The results are presented in Table \ref{table:Transformer}. We can find the following observations: 1) the Temporal Transformer can improve the performance of generalization capability. 2) more layers of Standard encoder for Temporal Transformer can not further improve the performance, which shows that one layer of standard Transformer encoder is sufficient in our framework.

\begin{table}[]
\footnotesize
\centering
\resizebox{1.0\columnwidth}{!}{
\tablestyle{7pt}{1.08}
\begin{tabular}{llllll}
\hline
\multirow{2}{*}{Model} & \multicolumn{4}{c}{Train on F2F} &      \\ \cline{2-5}
                       & DF     & FS     & F2F   & NT     & Avg  \\ \hline
3D R50 FTCN                   & 98.0   & 95.9   & \textbf{100}   & 96.0   & 97.5 \\
3D R50 FTCN+TT L\x1             & \textbf{98.1}   & \textbf{99.6}   & \textbf{100}   & \textbf{98.0}   & \textbf{98.9} \\
3D R50 FTCN+TT L\x2             & 97.8   & 98.6   & \textbf{100}   & 97.7   & 98.5 \\
3D R50 FTCN+TT L\x3             & 97.0   & 98.4   & \textbf{100}   & 97.4   & 98.2 \\ \hline
\end{tabular}
}
\vspace{0.5mm}
\caption{\textbf{Ablation study of using different layers of Transformer encoders in our framework.} Video-level AUC(\%) is reported.}
\label{table:Transformer}
\vspace{-3mm}
\end{table}

\vspace{-3mm}
\subsection{Localization of temporal incoherence}
\begin{figure} 
    \centering 
    \includegraphics[width=\linewidth]{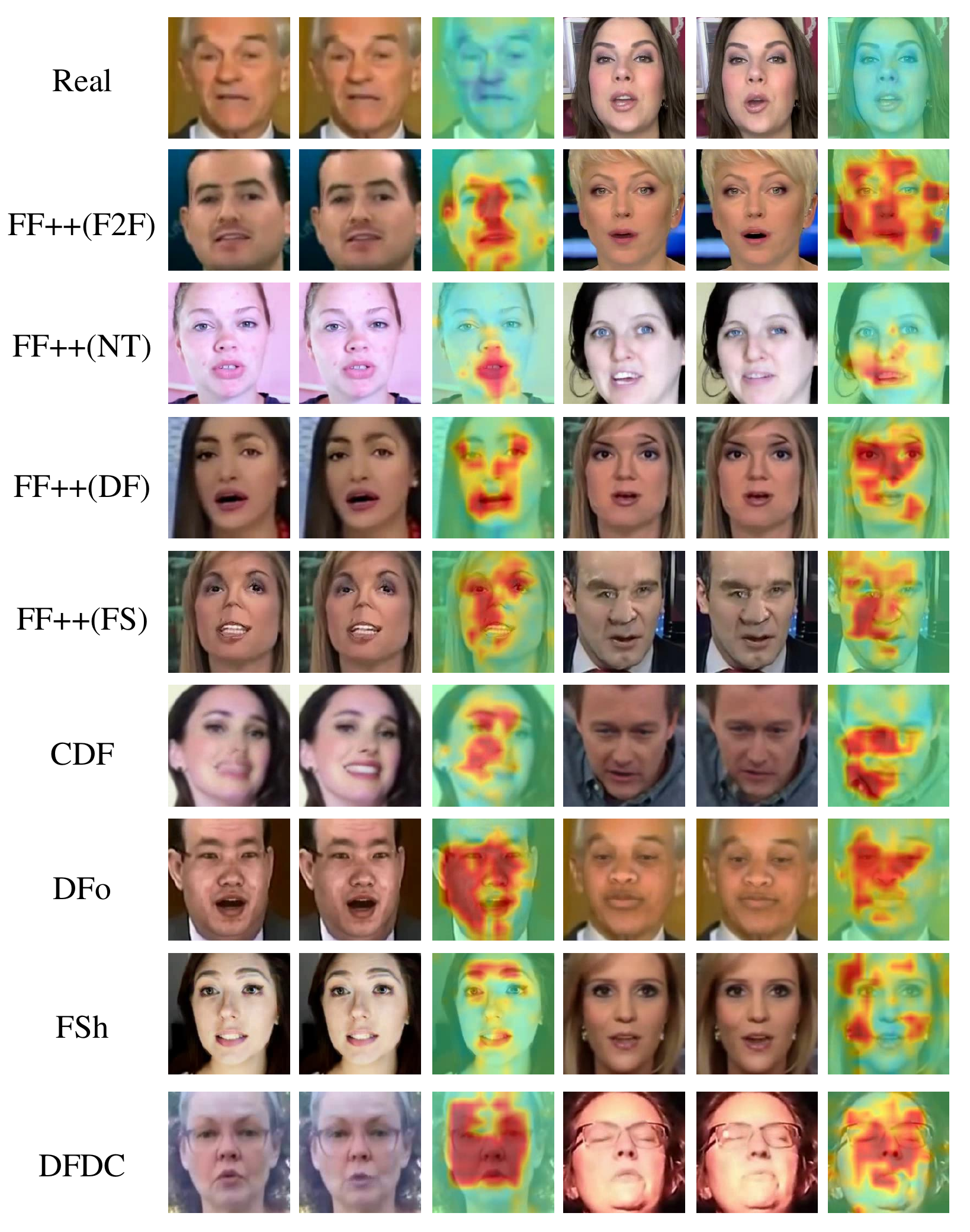}
    \caption{\textbf{Visualization of temporal defect localization on different datasets.} Each row shows two examples. For each example, the first two columns are consecutive frames in a video clip, the third column visualizes the localized defect regions, where warmer color indicates higher probability of forgery.}
    \label{fig:visualization}
    \vspace{-3mm}
\end{figure}

Without training with any explicit annotations, our method could be readily extended to localize temporally incoherent regions. In test time, for an input clip, we slide a window across the spatial domain. For regions outside the sliding window, we remove their content by replacing their RGB values with zeros. The modified clip is then fed into our forgery classifier to estimate the fake probability of the window area. Figure \ref{fig:visualization} shows that our method could robustly distinguish real and fake clips, and accurately localize the regions even with subtle temporal defects. For better visualization of temporal incoherence, please check the video results in the supplementary material. 

\vspace{-5mm}
\section{Conclusion}
\vspace{-2mm}
This paper investigates the effectiveness of temporal cues for more robust and general video face forgery detection. We propose to first encode short-term flickering with a Fully Temporal Convolution Network, then explore more subtle long-term incoherence with a Temporal Transformer. Extensive experiments evident the significant effects of the temporal information for video face forgery detection, and show the superior capabilities both on robustness and generalization of our proposed solution against previous methods. We hope our study will attract the community's attention to the temporal incoherence in deepfake detection.

\vspace{-3mm}
\section{Acknowledgements}
\vspace{-2mm}
Yinglin Zheng and Ming Zeng were partially supported by NSFC(No.62072382), Fundamental Research Funds for Central Universities, China(No.20720190003).

{\small
\bibliographystyle{ieee_fullname}
\bibliography{egbib}
}

\end{document}